\documentclass[10pt,twocolumn,letterpaper]{article}

\usepackage{cvpr}            % camera-ready
\usepackage{amsmath,amssymb,mathtools,bm}
\usepackage[normalem]{ulem}
\usepackage{xcolor}
\usepackage{microtype}
\usepackage{cuted}     % strip environment (official 2-column teaser support)

\usepackage{multirow}
\usepackage{xspace}
\newcommand{\methodname}{GENIQ\xspace}   % GENerated-Image Query (user, 2026-10-06); replaces provisional GIQ

\definecolor{cvprblue}{rgb}{0.21,0.49,0.74}
\usepackage[pagebackref,breaklinks,colorlinks,allcolors=cvprblue]{hyperref}

\def\paperID{*****} % OpenReview paper ID, after abstract registration (2026-11-10)
\def\confName{CVPR}
\def\confYear{2027}
\def\arxivversion{} % arXiv-only float placement in the shared sections (Figure 3, see 03_method.tex)

\title{Segment What It Can Draw: Querying a Text-to-Image Generator\\
  with Its Own Images for Skin Lesion Segmentation}

\author{
Hongyu Cao\textsuperscript{1}
\qquad Guangfu Guo\textsuperscript{1}
\qquad Jinghan Zhang\textsuperscript{1}
\\
Dakshak Keerthi Chandra\textsuperscript{2}
\qquad Ying Zhang\textsuperscript{3}
\qquad Kunpeng Liu\textsuperscript{1}
\\[6pt]
\textsuperscript{1}Clemson University, Clemson, SC, USA
\qquad
\textsuperscript{2}Analog Devices, Inc., USA
\\
\textsuperscript{3}Wake Forest University, USA
\\[2pt]
{\tt\small
hcao2@clemson.edu \quad
ggou@clemson.edu \quad
jinghaz@clemson.edu}
\\
{\tt\small
dakshak.keerthichandra@analog.com \quad
ying.zhang@wfu.edu \quad
kunpenl@clemson.edu
}
}
\begin{document}
\maketitle

% REVISION R (2026-10-13): text-specified target, set-level refinement, label-free tau; numbers from results/lineB/main/ci_means.json (ours).
% Tier 3 prose for the Abstract (REFRAME T, task first, 2026-10-11): four moves (task, resource + obstacle, method,
% evidence; the refinement-pool size, prompt dependence, and the position-invariance condition moved to the intro and RQ4, 2026-10-20); \methodname once per mention, no step names, no forward references, no citations;
% "training-free" only inside the constraint clause.
\begin{abstract}
Skin lesion segmentation without annotated masks requires the target to be specified through other cues. Existing approaches rely on dermoscopy-specific rules, annotation-derived model priors, or user-selected regions. A general text-to-image diffusion transformer offers a text interface to the target, but its concept-word attention is unreliable on dermoscopy images: in the model's attention-output space, the word-derived skin-to-lesion direction is poorly aligned with the direction separating real skin from real lesions. We show that the direction derived from generated healthy-skin and lesion images is substantially better aligned. Based on this finding, we propose \methodname{}, a training-free method that constructs initial skin and lesion prototypes from generated images and refines them using the most skin-like and lesion-like patches from unlabelled dermoscopy images. It scores test patches along the refined direction and applies a gated threshold to suppress contextual spill-over around small lesions. Without inference-time annotations or additional models, \methodname{} achieves Dice scores of 0.85--0.88 on three dermoscopy test sets. It outperforms zero-shot and word-based attention methods on all three sets, and a hand-designed dermoscopy pipeline in Dice and IoU on two.
\end{abstract}

% REVISION R (2026-10-13): P2 question, P3 category 3, P4 refinement, P5 contributions 1-2 updated.
% Tier 3 prose for §1 Introduction (REFRAME T, task first, 2026-10-11): task first; one challenge (word misalignment)
% whose diagnosis is stated here with the teaser; the bold prior-work list is kept for now (user, 2026-10-11) and
% reordered task first; one method name (\methodname); "generated images"; no "knows"; "training-free" never frames
% the paper. Teaser: home of [fx_1]; panel (b) = [c3].
\begin{figure*}[t]
  \centering
  \includegraphics[width=\textwidth]{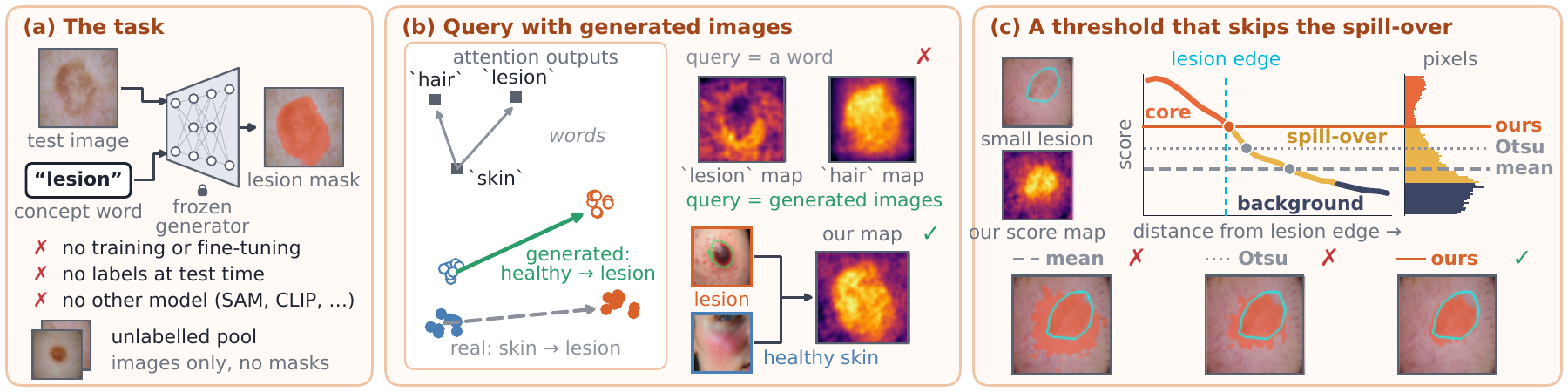}
  \caption{\textbf{GENIQ: task, query construction, and thresholding.}
(a) Given a dermoscopy image, a text-named target, frozen FLUX~\cite{labs2025flux1kontextflowmatching,flux2024}, and unlabelled images, GENIQ predicts a lesion mask without training, annotations, or external models.
(b) Concept-word directions align poorly with the real skin-to-lesion direction, whereas the generated healthy-skin-to-lesion direction aligns better. GENIQ uses the word only inside generated images and refines the resulting query on unlabelled real images (green).
(c) On a small lesion, the mean and two-class Otsu thresholds keep the score spill-over, whereas the upper threshold of a three-level split lies near the lesion edge. Cyan: clinician’s outline; histogram colours: the three score levels.}
\label{fig:teaser}
\end{figure*}

\section{Introduction}\label{sec:intro}

Skin lesion segmentation delineates lesions in dermoscopy images, enabling the measurement of their size, border, and change over time~\cite{zeng2023unsupervisedskinlesionsegmentation, ramella2021saliencybasedsegmentationdermoscopicimages}. 
However, obtaining pixel-level masks is costly, motivating segmentation methods that require no mask annotations during training or inference. Without such annotations, a central challenge is how to specify the target region to be segmented. Existing approaches rely on dermoscopy-specific rules, annotation-derived priors embedded in pretrained models, or manual region selection. 
Text-to-image generators offer another potential source of guidance: when prompted with a lesion-related word, they can generate both lesions and surrounding healthy skin, suggesting that their internal features may encode the distinction between the two.

% Skin lesion segmentation outlines lesions in dermoscopy images for measuring their size, border, and change over time~\cite{zeng2023unsupervisedskinlesionsegmentation, ramella2021saliencybasedsegmentationdermoscopicimages}. 
% Because pixel-level masks are costly to draw, methods have been evaluated without them on datasets including ISIC-2018~\cite{codella2019skinlesionanalysismelanoma}, PH2~\cite{PH2-6610779}, and HAM10000~\cite{Tschandl_2018}. Without masks, the target must be specified by other means: dermoscopy-specific rules, annotations carried in through pretrained models, or a person selecting a region. A general text-to-image generator offers another source of guidance: given a word, it draws both lesions and healthy skin, suggesting that its features may distinguish them.

% ConceptAttention~\cite{helbling2025conceptattentiondiffusiontransformerslearn} provides a way to read these features for segmentation without training. A diffusion transformer represents image patches as tokens whose attention outputs describe each patch in the context of the image~\cite{tang2022daaminterpretingstablediffusion}. ConceptAttention passes a concept word through the same layers as an extra token and scores each patch by its proximity to that word relative to other words in the shared attention-output space.

Turning this observation into segmentation requires a mechanism for reading the generator's internal features at the patch level. ConceptAttention~\cite{helbling2025conceptattentiondiffusiontransformerslearn} provides such a mechanism without additional training. In a diffusion transformer, image patches are represented as tokens whose attention outputs encode each patch in the context of the entire image~\cite{tang2022daaminterpretingstablediffusion}. 
To localize a target, ConceptAttention introduces its concept word as an additional token, processes it through the same attention layers, and scores each image patch by its proximity to the concept token relative to other words in the shared attention-output space. 
It thus converts a text-named target into a training-free, patch-level word query.

% On dermoscopy images, however, this word query is unreliable.
% In FLUX~\cite{labs2025flux1kontextflowmatching, flux2024}, the \texttt{lesion} word map ranks lesion patches no better than the mean ground-truth mask of other images; the \texttt{hair} map can even highlight a mole in an image without hair (Figure~\ref{fig:teaser}b, top right). 
% The attention outputs themselves remain useful: the direction from \texttt{skin} to \texttt{lesion} has a cosine of only 0.21 with the real skin-to-lesion direction, about as low as \texttt{skin} to \texttt{hair}, whereas the direction from generated healthy skin to generated lesions reaches 0.61 (Figure~\ref{fig:teaser}b, left). 
% How can we query these features using only a text-named target, without training, labels at inference, or additional models (Figure~\ref{fig:teaser}a)?

On dermoscopy images, however, this word-based query is unreliable. In FLUX~\cite{labs2025flux1kontextflowmatching, flux2024}, the \texttt{lesion} word map ranks lesion patches no better than a spatial prior obtained by averaging the ground-truth masks of other images. The failure is not limited to lesion-related words: the \texttt{hair} map can even highlight a mole in an image containing no hair (Figure~\ref{fig:teaser}b, top right).
% reason analysis
These failures appear to arise from the query rather than from the image-token features themselves. 
In the shared attention-output space, the vector from the \texttt{skin} word to the \texttt{lesion} word has a cosine similarity of only 0.21 with the direction from real skin patches to real lesion patches. This alignment is approximately as weak as that of the \texttt{skin}-to-\texttt{hair} word vector. 
By contrast, the direction from generated healthy-skin patches to generated lesion patches achieves a cosine similarity of 0.61 with the real skin-to-lesion direction (Figure~\ref{fig:teaser}b, left). Thus, the feature space contains a useful skin-to-lesion direction, but concept words do not reliably recover it.
This raises our central question: can we use a target specified only by text to construct a reliable query for these features, without training, inference-time annotations, or external models (Figure~\ref{fig:teaser}a)?

Prior work addresses complementary parts of this problem but leaves a gap.
\textbf{1) Label-free lesion segmentation} uses dermoscopy-specific pipelines or pretrained models, including superpixel clustering after hair removal, SAM prompted by vision--language saliency, and Stable Diffusion self-attention grouped into regions~\cite{koleilat2025medclipsamv2universaltextdrivenmedical, hamrani2025selfattentiondiffusionmodelszeroshot, kirillov2023segment, tian2024diffuseattendsegmentunsupervised}. However, the target is still specified through domain-specific rules, annotation-derived model priors, or human selection, making these methods difficult to extend to new text-named findings.
\textbf{2) Word read-outs of diffusion models} localise objects by scoring image patches against prompt or concept words~\cite{kim2025seg4diffunveilingopenvocabularysegmentation, zhang2026anchordifftrainingfreeconceptgrounding}. Because the concept word remains the query, even when prototypes from unlabelled images are introduced~\cite{snell2017prototypicalnetworksfewshotlearning}, these methods retain the word-to-image misalignment observed above.
\textbf{3) Generated references} construct visual prototypes from generated class images, using class-word cross-attention to separate foreground and background patches~\cite{karazija2024diffusionmodelsopenvocabularysegmentation, barsellotti2024trainingfreeopenvocabularysegmentationoffline}. However, they typically use background patches from the same class image rather than an explicitly generated negative class, such as healthy skin, and often extract features with an external encoder. They therefore do not provide the healthy-skin-to-lesion direction in the generator's own attention-output space.
Together, these methods do not combine text-based target specification with an image-derived query constructed in the generator's own attention-output space.

\textbf{Our Approach: Querying with Generated Images.}
To fill this gap, we propose \methodname{}, which replaces ConceptAttention's direct word query with an image-derived query constructed by the same generator. 
Rather than using the concept word to match real lesion patches directly, we use it only within generated images, where distinguishing a generated lesion from its surrounding skin is easier. 
Generated lesion and healthy-skin images then define a visual direction for querying real dermoscopy images (Figure~\ref{fig:teaser}b, right). 
We further adapt this direction using unlabelled real images and introduce a gated threshold to reduce spill-over onto skin around small lesions (Figure~\ref{fig:teaser}c).
In this way, generated images bridge the gap between a text-named target and its visual appearance in real images.

Our contributions are as follows:
1) We introduce a generated-image query that replaces the poorly aligned concept-word query with a healthy-skin-to-lesion direction constructed in the generator's own attention-output space.
2) We refine this query using unlabelled dermoscopy images, adapting it to real lesion appearance without training or annotations.
3) We introduce a gated threshold that reduces attention spill-over around small lesions.
4) On three dermoscopy test sets, \methodname{} outperforms zero-shot baselines and word-based read-outs, and surpasses a hand-designed dermoscopy pipeline on two of the three sets.
% REVISION R (2026-10-13): MaViSeg-style visual prototypes are additive to word scores; test-time prototypes need a starting query (ours: generated images).
% Tier 3 prose for Related Work (paper/semantic.md §2, REFRAME T 2026-10-11: moved to §2, task-first order). User rule (2026-10-04):
% prose only, no \cite; candidate references as comments (bib keys from docs/references.bib where present; ids marked "verify"
% are unchecked). Third person only, bold on the four category labels only, no closer.
\section{Related Work}\label{sec:related}

% 2026-10-21 (user): Figure 2 (the method figure, Sec. 3) is defined here so that it prints one page earlier, on page 3.
\begin{figure*}[t]
  \centering
  % 2026-10-11 trial: draft "B aligned" (trunk + three change cards); the old figure is kept below, commented out.
  % \includegraphics[width=\textwidth]{figures/fig_framework.pdf}
  % \caption{\methodname{}. The generator produces lesion and healthy-skin images, and the concept word only marks the lesion inside each generated lesion image (Step~1). Each patch of a real image is scored along the direction from the healthy-skin prototype to the lesion prototype, in the space of attention outputs where ConceptAttention scores concept words (Step~2). A gate on how well the map separates into levels decides how the map is thresholded (Step~3).}
  % REVISION R (2026-10-13): four-card figure (paper_figs.py fig_method4); the three-card fig_method.pdf and its caption stay in git (tag paper-before-I1).
  \includegraphics[width=\textwidth]{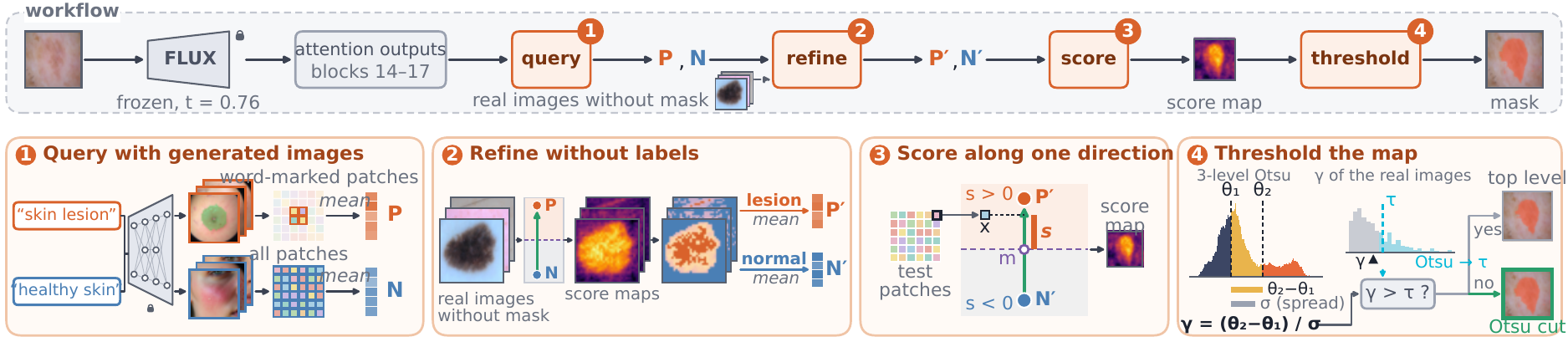}
  % 2026-10-19 (user): steps in the order they run (query, refine, score, threshold); old caption in figures_backup/2026-10-19_before_reorder.
  % 2026-10-21 (user): caption compacted (8 -> about 6 lines); dropped what the figure labels (the step outputs, the
  % lesion/normal labels of card 2, gamma = (theta2 - theta1)/sigma of card 4).
  % TODO：加一个直角线对应workflow中的block和下面的block
  % 2026-11-03 (page compression): per-step details left to Secs. 3.3-3.6; what each step passes on is kept.
  \caption{\methodname{}. Top: the workflow; its grey parts (FLUX, attention outputs) are kept from ConceptAttention, and the real images without masks (dashed) are read once before testing, by Step~2 and, for $\tau$, by Step~4. Bottom: the four steps (Sections~\ref{sec:method-proto}--\ref{sec:method-gap}) and what each passes on. Step~1: $\mathbf{P},\mathbf{N}$ from generated images. Step~2: $\mathbf{P}',\mathbf{N}'$ from the top 20\% and bottom 50\% of the real images' score maps. Step~3: the map from each patch's signed distance to $\mathbf{m}$ along $\mathbf{N}'\to\mathbf{P}'$. Step~4: the top level if $\gamma>\tau$, else one Otsu cut (as here).}
  \label{fig:framework}
\end{figure*}

\textbf{Label-free inference for medical image segmentation.}
Classical dermoscopy pipelines remove hair, correct colour, and cluster superpixels or saliency regions without annotations~\cite{zeng2023unsupervisedskinlesionsegmentation, ramella2021saliencybasedsegmentationdermoscopicimages}. Diffusion-based approaches either group Stable Diffusion self-attention into candidate regions for a clinician to select, or learn medical adapters from annotated data~\cite{tian2024diffuseattendsegmentunsupervised, hamrani2025selfattentiondiffusionmodelszeroshot, jiang2025advancinggeneralizabletumorsegmentation, vilouras2025zeroshotmedicalphrasegrounding}. Other methods use BiomedCLIP saliency to prompt SAM~\cite{koleilat2025medclipsamv2universaltextdrivenmedical}, or replace text prompts with prototypes from labelled real images~\cite{tang2025prototypebasedimagepromptingweakly}. Thus, although these methods may require no annotation for the test image, their target specification is supplied by domain-specific rules, human selection, labelled examples, or medically supervised models. Our setting instead specifies the target by text interpreted by a general-purpose generator without medical-specific training.
\textbf{Concept read-outs from diffusion models.}
Diffusion models expose spatial information about prompt concepts through their internal attention. In UNet-based models, cross-attention between a word and image positions can be aggregated into a heat map, which training-free segmenters subsequently filter and refine~\cite{tang2022daaminterpretingstablediffusion, wang2024diffusionmodelsecretlytrainingfree}. In multimodal diffusion transformers, ConceptAttention~\cite{helbling2025conceptattentiondiffusiontransformerslearn} introduces the concept word as an additional token and compares its attention output with those of image patches. Later methods add grounding layers, propagate scores between patches, or incorporate visual prototypes constructed from unlabelled real images~\cite{kim2025seg4diffunveilingopenvocabularysegmentation, kukleva2026refamattentionmagnetszeroshot, zhang2026anchordifftrainingfreeconceptgrounding, chakraborty2026mavisegmanifoldpropagationvisual, jun2026interpretablemotionattentivemapsspatiotemporally}. These refinements retain the concept word as the query or as the anchor for the final score. Moreover, they have primarily been evaluated on natural-image categories, leaving settings with strong word-to-visual misalignment largely unexplored.
\textbf{Generated references for segmentation.}
Generated-reference methods use synthetic class images to construct visual prototypes. OVDiff~\cite{karazija2024diffusionmodelsopenvocabularysegmentation} localises objects in Stable Diffusion images through cross-attention and forms class, part, and background prototypes; one variant encodes both generated and test images using the generator's own features. FreeDA~\cite{barsellotti2024trainingfreeopenvocabularysegmentationoffline} and FOSSIL~\cite{Barsellotti_2024_WACV} similarly construct generated references but encode them with an external vision model. Other methods form prototypes directly from unlabelled test images~\cite{ma2026perceptualanchoringprototypeguidedtext}, while generated image--mask pairs can instead be used to train a segmenter~\cite{wu2024diffumasksynthesizingimagespixellevel, mao2025medsegfactorytextguidedgenerationmedical}. 
These approaches do not construct an explicit negative-to-target direction from generated classes in the generator's attention-output space: generated-reference methods use background patches from target-class images, whereas test-time prototypes require an existing query to identify target patches. 

\section{Method}\label{sec:method}

\subsection{Overview}\label{sec:method-overview}

\methodname{} turns a text-named target into a segmentation mask in four steps (Figure~\ref{fig:framework}). 
Step~1 translates the target word into a visual query: the generator produces lesion and healthy-skin images, and the concept word identifies lesion patches only within the generated images. 
The resulting lesion and skin prototypes define an initial healthy-skin-to-lesion direction. 
Step~2 bridges the synthetic-to-real gap by using this direction to find lesion-like and skin-like patches in an unlabelled dermoscopy pool, which refine the two prototypes before testing. 
Step~3 projects each test patch onto the refined direction to produce a lesion score map. 
Step~4 converts the score map into a mask with a gated threshold that separates small lesions from attention spill-over (Section~\ref{sec:method-gap}).
\methodname{} uses the same image-token attention outputs as ConceptAttention, but replaces its direct word query with the refined image-derived direction and its map-mean threshold with the gated threshold.

\subsection{Task and Preliminaries}\label{sec:pf}\label{sec:prelim}

\textbf{Task.}
Given a dermoscopy image
$I\in\mathbb{R}^{H\times W\times 3}$,
a frozen text-to-image diffusion transformer $G$,
a textual target specification $\mathcal{T}$,
and a pool $\mathcal{U}$ of unlabelled dermoscopy images, the task is to predict a binary lesion mask
$M\in\{0,1\}^{H\times W}$,
where $H$ and $W$ are its height and width.

The textual specification is
$\mathcal{T}=(w,\mathcal{C},\mathcal{D}^{+},\mathcal{D}^{-})$,
where $w$ is the target concept;
$\mathcal{C}\ni w$ is a list of concepts that may appear in the image;
and $\mathcal{D}^{+}$ and $\mathcal{D}^{-}$ are prompts describing the target and its healthy reference, respectively.
Here $w=\texttt{lesion}$ and $\mathcal{C}=\{\texttt{lesion},\texttt{skin},\texttt{hair},\texttt{ruler marking}\}$; the prompts are given in Supp.~C.
Only $\mathcal{T}$ specifies the target: $\mathcal{U}$ contains no masks or target identities.
The method updates no parameter, uses no annotation during inference, and invokes no model other than $G$.
Clinician masks are used only for evaluation.

\textbf{Image-Token Attention Outputs.}
We represent an image using attention outputs from the double-stream blocks of a multimodal diffusion transformer such as FLUX~\cite{labs2025flux1kontextflowmatching, flux2024}.
The image is encoded by the model's autoencoder, perturbed at a fixed diffusion timestep $t$, and passed through $G$ once with a fixed caption.
Let $\mathbf{o}^{(\ell)}_p$ denote the $2\times2$ spatially pooled attention output of patch $p$ in block $\ell$, taken before the attention output projection.
Averaging over a set of blocks $\mathcal{L}$ gives
\begin{equation}\label{eq:features}
  \mathbf{x}_p
  =
  \frac{1}{|\mathcal{L}|}
  \sum_{\ell\in\mathcal{L}}
  \mathbf{o}^{(\ell)}_p,
  \qquad p\in\Omega,
\end{equation}
where $\Omega$ is the pooled patch grid and $\mathbf{x}_p$ is the representation of patch $p$ used in the following subsections.

\textbf{Concept-Word Marking.}
Following ConceptAttention~\cite{helbling2025conceptattentiondiffusiontransformerslearn}, each concept $c\in\mathcal{C}$ is introduced as an additional concept token.
Let $\mathbf{z}^{(\ell)}_c$ denote its attention output in block $\ell$.
Concept tokens attend to image tokens, but image tokens do not attend back, so $\mathbf{x}_p$ is independent of $\mathcal{C}$.
The soft weight of concept $c$ at patch $p$ is
\begin{equation}\label{eq:conceptattention}
  A_{c,p}
  =
  \frac{1}{|\mathcal{L}|}
  \sum_{\ell\in\mathcal{L}}
  \frac{
    \exp\!\big(
      \mathbf{o}^{(\ell)\top}_p
      \mathbf{z}^{(\ell)}_c
    \big)
  }{
    \sum_{c'\in\mathcal{C}}
    \exp\!\big(
      \mathbf{o}^{(\ell)\top}_p
      \mathbf{z}^{(\ell)}_{c'}
    \big)
  }.
\end{equation}
We call $A_{w,\cdot}$ the target concept map.
\methodname{} uses this map only to mark the target in generated images; real-image scores are defined from $\mathbf{x}_p$ in the following subsections.

\subsection{Querying with Generated Images}\label{sec:method-proto}

\textbf{Why Generated Images Define the Query.}
As observed in Figure~\ref{fig:teaser}b, the direction from the \texttt{skin} concept token to the \texttt{lesion} concept token is poorly aligned with the visual direction from real skin to real lesions, whereas the direction derived from generated images is better aligned (Section~\ref{sec:rq-why}).
We therefore use the concept word only to locate lesions within generated lesion images, where it need only distinguish a known lesion from the surrounding skin (Supp.~C shows the marked regions).
The query is built from the image-token outputs of these generated lesions and of generated healthy skin.
\textbf{Assumption 1 (Generated-Direction Fidelity).}
The direction between generated healthy-skin and lesion features is sufficiently aligned with the corresponding direction in real dermoscopy images to provide an initial query.
Section~\ref{sec:rq-why} tests it.

\textbf{Prototype Construction.}
The generator $G$ samples a set $\mathcal{E}^{+}$ of lesion images from the target prompts $\mathcal{D}^{+}$ and a set $\mathcal{E}^{-}$ of healthy-skin images from the healthy-reference prompts $\mathcal{D}^{-}$.
For a generated image $e$, let $\mathbf{x}^{(e)}_p$ denote the representation of patch $p$ from Equation~\eqref{eq:features}, and let $A^{(e)}_{c,p}$ denote its concept weight from Equation~\eqref{eq:conceptattention}.
For each lesion image $e\in\mathcal{E}^{+}$, we define its marked lesion region as
\begin{equation}\label{eq:generated-region}
  R_e
  =
  \left\{
    p\in\Omega:
    \arg\max_{c\in\mathcal{C}} A^{(e)}_{c,p}=w
  \right\},
\end{equation}
where $\Omega$ is the patch grid, $\mathcal{C}$ is the concept list, and $w$ is the target concept.

The generated lesion prototype $\mathbf{P}$ and healthy-skin prototype $\mathbf{N}$ are
\begin{equation}\label{eq:prototypes}
\begin{aligned}
  \mathbf{P}
  &=
  \frac{1}{|\mathcal{E}^{+}|}
  \sum_{e\in\mathcal{E}^{+}}
  \frac{1}{|R_e|}
  \sum_{p\in R_e}
  \mathbf{x}^{(e)}_p,\\
  \mathbf{N}
  &=
  \frac{1}{|\mathcal{E}^{-}|}
  \sum_{e\in\mathcal{E}^{-}}
  \frac{1}{|\Omega|}
  \sum_{p\in\Omega}
  \mathbf{x}^{(e)}_p.
\end{aligned}
\end{equation}
Averaging within each image first gives every generated image equal weight, regardless of lesion size.

The difference between $\mathbf{P}$ and $\mathbf{N}$ defines the initial generated-image query.
We generate $|\mathcal{E}^{+}|=|\mathcal{E}^{-}|=16$ reference images once, using two prompts and eight random seeds for each class; further details are in Supp.~C.

\begin{table*}[t]
\centering
\small
\setlength{\tabcolsep}{4pt}
\caption{Lesion masks on three dermoscopy test sets. Dice and IoU at $224{\times}224$, mean with the half-width of the 95\% bootstrap confidence interval over test images as a subscript; best mean per column in bold; paired differences in Table~S6. Map-producing baselines are thresholded by the rule of Section~\ref{sec:method-gap} with their own $\tau$ fitted on the labelled validation split, ours with the label-free $\tau$ of its unlabelled pool; mask-producing methods use their own rule. Rotated labels and parentheses give the baseline groups of Section~\ref{sec:setup}; FLUX and Other models split the baselines evaluated by ConceptAttention by backbone, and Control keeps our generated images and regions but swaps the feature extractor. $^{\dagger}$Strongest of the seven Chefer-codebase CLIP explainability variants evaluated by ConceptAttention; all seven, the remaining hand-designed methods, and paired differences are in the supplementary material. $^{\ddagger}$Official DiffSeg code with ADZUS's settings; the lesion is the region nearest the image centre, a label-free rule chosen on the validation split (Table~S4).}
\label{tab:main}
\newcommand{\grp}[2]{\multirow{#1}{*}{\rotatebox[origin=c]{90}{\scriptsize\itshape\begin{tabular}{@{}c@{}}#2\end{tabular}}}}
\begin{tabular}{@{}c@{\hspace{3pt}}lcccccc}
\toprule
 & & \multicolumn{2}{c}{ISIC-2018 (1{,}000)} & \multicolumn{2}{c}{PH2 (200)} & \multicolumn{2}{c}{HAM10000 (1{,}000)} \\
\cmidrule(lr){3-4}\cmidrule(lr){5-6}\cmidrule(lr){7-8}
\multicolumn{2}{l}{Method} & Dice & IoU & Dice & IoU & Dice & IoU \\
\midrule
\multicolumn{2}{l}{Mean-mask prior (no image)} & 0.620$_{\pm 0.012}$ & 0.474$_{\pm 0.011}$ & 0.642$_{\pm 0.019}$ & 0.489$_{\pm 0.021}$ & 0.640$_{\pm 0.011}$ & 0.495$_{\pm 0.011}$ \\
\cmidrule(l){1-8}
\grp{2}{Zero-\\shot} & MedCLIP-SAMv2 & 0.463$_{\pm 0.023}$ & 0.384$_{\pm 0.022}$ & 0.362$_{\pm 0.046}$ & 0.289$_{\pm 0.043}$ & 0.508$_{\pm 0.021}$ & 0.413$_{\pm 0.019}$ \\
 & ADZUS (DiffSeg)$^{\ddagger}$ & 0.536$_{\pm 0.017}$ & 0.421$_{\pm 0.018}$ & 0.524$_{\pm 0.035}$ & 0.403$_{\pm 0.036}$ & 0.458$_{\pm 0.015}$ & 0.335$_{\pm 0.015}$ \\
\cmidrule(l){1-8}
\grp{3}{FLUX} & ConceptAttention, layers 14--17 & 0.549$_{\pm 0.018}$ & 0.431$_{\pm 0.017}$ & 0.694$_{\pm 0.026}$ & 0.562$_{\pm 0.028}$ & 0.683$_{\pm 0.016}$ & 0.570$_{\pm 0.016}$ \\
 & ConceptAttention, layers 9--13 & 0.706$_{\pm 0.013}$ & 0.579$_{\pm 0.014}$ & 0.769$_{\pm 0.024}$ & 0.654$_{\pm 0.028}$ & 0.785$_{\pm 0.012}$ & 0.680$_{\pm 0.013}$ \\
 & FLUX cross-attention & 0.612$_{\pm 0.012}$ & 0.466$_{\pm 0.012}$ & 0.694$_{\pm 0.020}$ & 0.550$_{\pm 0.022}$ & 0.732$_{\pm 0.013}$ & 0.615$_{\pm 0.014}$ \\
\cmidrule(l){1-8}
\grp{4}{Other models} & DAAM (SD-XL) & 0.369$_{\pm 0.011}$ & 0.240$_{\pm 0.008}$ & 0.416$_{\pm 0.019}$ & 0.273$_{\pm 0.016}$ & 0.414$_{\pm 0.010}$ & 0.275$_{\pm 0.008}$ \\
 & CLIP TextSpan & 0.575$_{\pm 0.009}$ & 0.416$_{\pm 0.008}$ & 0.554$_{\pm 0.019}$ & 0.395$_{\pm 0.017}$ & 0.618$_{\pm 0.009}$ & 0.462$_{\pm 0.009}$ \\
 & DINO self-attention & 0.534$_{\pm 0.016}$ & 0.406$_{\pm 0.015}$ & 0.494$_{\pm 0.028}$ & 0.349$_{\pm 0.023}$ & 0.560$_{\pm 0.015}$ & 0.427$_{\pm 0.014}$ \\
 & CLIP rollout$^{\dagger}$ & 0.539$_{\pm 0.014}$ & 0.400$_{\pm 0.012}$ & 0.489$_{\pm 0.035}$ & 0.357$_{\pm 0.029}$ & 0.556$_{\pm 0.014}$ & 0.414$_{\pm 0.013}$ \\
\cmidrule(l){1-8}
\grp{3}{Control} & BiomedCLIP features & 0.718$_{\pm 0.011}$ & 0.587$_{\pm 0.012}$ & 0.783$_{\pm 0.017}$ & 0.658$_{\pm 0.021}$ & 0.773$_{\pm 0.009}$ & 0.651$_{\pm 0.011}$ \\
 & DINOv2 features & 0.729$_{\pm 0.009}$ & 0.593$_{\pm 0.010}$ & 0.790$_{\pm 0.014}$ & 0.664$_{\pm 0.018}$ & 0.757$_{\pm 0.008}$ & 0.625$_{\pm 0.009}$ \\
 & CLIP features & 0.423$_{\pm 0.014}$ & 0.297$_{\pm 0.013}$ & 0.467$_{\pm 0.030}$ & 0.334$_{\pm 0.028}$ & 0.425$_{\pm 0.013}$ & 0.294$_{\pm 0.011}$ \\
\cmidrule(l){1-8}
\multicolumn{2}{l}{SLED (hand-designed)} & 0.779$_{\pm 0.016}$ & 0.697$_{\pm 0.017}$ & \textbf{0.903}$_{\pm 0.013}$ & \textbf{0.835}$_{\pm 0.018}$ & 0.822$_{\pm 0.013}$ & 0.741$_{\pm 0.014}$ \\
\midrule
\multicolumn{2}{l}{\textbf{Ours}} & \textbf{0.845}$_{\pm 0.009}$ & \textbf{0.752}$_{\pm 0.011}$ & 0.877$_{\pm 0.012}$ & 0.790$_{\pm 0.017}$ & \textbf{0.859}$_{\pm 0.007}$ & \textbf{0.766}$_{\pm 0.009}$ \\
\bottomrule
\end{tabular}
\end{table*}

\subsection{Refining with Ranked Real Patches}\label{sec:method-refine}

\textbf{Why Refine with Unlabelled Images.}
The generated prototypes $\mathbf{P}$ and $\mathbf{N}$ identify the visual direction from healthy skin to lesion, but their endpoints reflect synthetic rather than real dermoscopy appearance (Supp.~C).
We therefore use the generated direction to rank the patches of the unlabelled pool $\mathcal{U}$ and replace the prototypes with the most lesion-like and skin-like real patches.
The generated direction thus supplies the target identity, and the selected patches supply the real-domain appearance; no mask or spatial prior is needed.

\textbf{Refined Prototypes.}
% 2026-11-03 (page compression): one score function for Steps 2 and 3 (former Eqs. rank and discriminant).
For a patch representation $\mathbf{x}$ and two prototypes $\mathbf{P},\mathbf{N}$, we define the score
\begin{equation}\label{eq:discriminant}
  s(\mathbf{x};\mathbf{P},\mathbf{N})
  =
  \left(
    \mathbf{x}
    -
    \frac{\mathbf{P}+\mathbf{N}}{2}
  \right)^{\!\top}
  (\mathbf{P}-\mathbf{N}),
\end{equation}
the position of $\mathbf{x}$ along the direction from $\mathbf{N}$ to $\mathbf{P}$; larger values are more lesion-like.
Let $T_u$ contain the top $q_{\mathrm{hi}}$ fraction of the patches of an unlabelled image $u\in\mathcal{U}$ by $s(\mathbf{x}^{(u)}_p;\mathbf{P},\mathbf{N})$, and let $B_u$ contain the bottom $q_{\mathrm{lo}}$ fraction.
Pooling the selected patches over the unlabelled images gives the refined lesion and skin prototypes
\begin{equation}\label{eq:refine}
\begin{aligned}
  \mathbf{P}'
  &=
  \operatorname{mean}
  \left\{
    \mathbf{x}^{(u)}_p
    :
    u\in\mathcal{U},\ p\in T_u
  \right\},\\
  \mathbf{N}'
  &=
  \operatorname{mean}
  \left\{
    \mathbf{x}^{(u)}_p
    :
    u\in\mathcal{U},\ p\in B_u
  \right\}.
\end{aligned}
\end{equation}
We use $q_{\mathrm{hi}}=0.2$ and $q_{\mathrm{lo}}=0.5$.
The refinement runs once before inference, and its prototypes are shared by all test images; no test image contributes to them.
Section~\ref{sec:rq-sens} studies the effect of the pool size and patch fractions.

\subsection{Scoring along One Direction}\label{sec:method-map}

\textbf{Why Score along One Direction.}
We score each test patch by its position along the refined direction $\mathbf{P}'-\mathbf{N}'$.
Unlike the concept-word map in Equation~\eqref{eq:conceptattention}, this score involves neither competition with distractor concepts nor a classifier fitted from labelled data.
It does, however, assume that the same lesion-to-skin contrast applies across spatial locations.
\textbf{Assumption 2 (Position-Invariant Contrast).}
The target tends to differ from its surroundings in the same feature-space direction wherever it appears in the image.
We adopt this assumption for dermoscopy, where lesions generally differ from surrounding skin in colour and texture regardless of position.

\textbf{Lesion Score.}
Each test patch $p\in\Omega$ is scored with the refined prototypes, $s_p=s(\mathbf{x}_p;\mathbf{P}',\mathbf{N}')$.
A larger $s_p$ indicates a more lesion-like patch, and
$s_p>0$ exactly when $\mathbf{x}_p$ is closer to $\mathbf{P}'$ than to $\mathbf{N}'$ in Euclidean distance.
Thus, Equation~\eqref{eq:discriminant} is a nearest-centroid linear discriminant with identity covariance.
Neither $\mathbf{x}_p$ nor $\mathbf{P}'-\mathbf{N}'$ is normalised.

For efficiency, all patch representations and prototypes are mapped to 256 dimensions by the same fixed random projection before scoring; the projection is omitted from the notation and described in Supp.~A.
The patch scores $\{s_p:p\in\Omega\}$ are bilinearly upsampled to the image resolution to form the score map $S$.

\ifdefined\arxivversion% Figure 3, input from 04_experiments.tex (anonymous main.pdf) or from 03_method.tex (arXiv, \arxivversion defined).
\begin{figure*}[t]
  \centering
  \includegraphics[width=\textwidth]{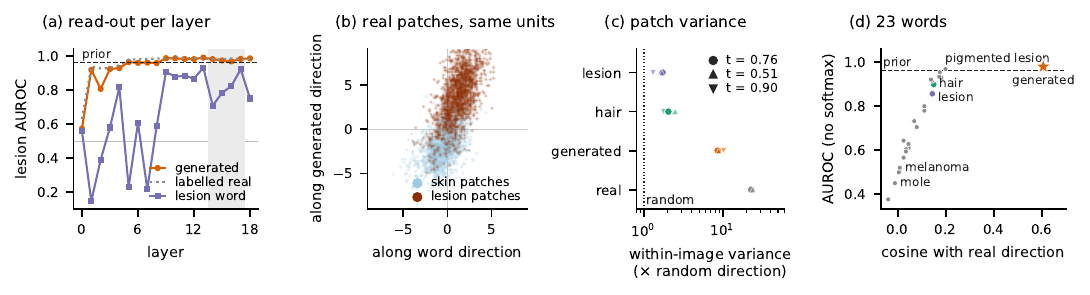}
  \caption{Why generated images work and the concept word fails (validation split; (b)--(d) at layers 14--17; dashed: mean-mask prior). (a) Lesion AUROC per double-stream layer of the word map, our map, and labelled real prototypes ($t=0.76$). (b) Patches minus their image mean along the unit word and generated directions (same units; 1{,}500 per class). (c) Within-image variance along each direction relative to a random one. (d) 23 words: cosine with the real skin-to-lesion direction against AUROC without softmax; star: generated direction.}
  \label{fig:why}
\end{figure*}
\fi
\subsection{Thresholding the Map by Level Separation}\label{sec:method-gap}

\textbf{Why Gate the Threshold on Separation.}
Because attention mixes information across tokens, patches near a lesion can inherit part of its context and receive intermediate scores.
The resulting map can therefore exhibit three approximate levels: lesion, contextual spill-over, and background (Figure~\ref{fig:teaser}c).
For a small lesion, a two-class threshold may group the spill-over with the lesion.
We therefore retain only the highest score level when it is clearly separated from the remaining map, and otherwise fall back to a two-class threshold.

\textbf{Gated Threshold.}
Let
$\theta_1<\theta_2$
be the two thresholds returned by three-class multi-Otsu,
$\mathrm{MultiOtsu}_3(S)$,
and let $\mathrm{Otsu}(S)$ denote the threshold returned by two-class Otsu~\cite{Otsu4310076}.
We measure the separation of the highest score level relative to the map's own spread, $\gamma=(\theta_2-\theta_1)/\sigma(S)$, where $\sigma(S)$ is the standard deviation of the pixel scores in $S$, so the gate does not depend on the scale of the scores.
The final mask is
\begin{equation}\label{eq:gap}
  M
  =
  \begin{cases}
    \mathbf{1}[S>\theta_2],
      & \text{if } \gamma>\tau,\\[2pt]
    \mathbf{1}[S>\mathrm{Otsu}(S)],
      & \text{otherwise},
  \end{cases}
\end{equation}
where $\tau$ is the minimum separation to keep only the top level.

Both $\gamma$ and $\tau$ are obtained without annotations.
For each image in the unlabelled pool $\mathcal{U}$, we construct a score map using $\mathbf{P}'$ and $\mathbf{N}'$ and compute its separation $\gamma$.
We then set $\tau$ to the two-class Otsu threshold of these separations.
Thus, only maps with a clearly isolated top level use $\theta_2$.

% REVISION R (2026-10-13): numbers of the refined method (results/lineB/review/RUNLOG.md, main/*.json); old prose under tag paper-before-I1.
% Tier 3 prose for §4 Experiments (REFRAME T, 2026-10-11): descriptive headings (2026-10-18, user: prefixed "RQn:" after a Research Questions subsection); comparison with
% label-free lesion segmenters first (SLED first), then why it works, threshold, design and sensitivity, case study;
% the labelled validation split gets one sentence here, details in Supp. app:valsplit (user 2026-10-11).
% Tight-page form (Setup / Findings; Question lines dropped and case study merged into 4.2 for the 8-page budget, 2026-10-11). Structural edits go through Tier 1/2 first.
\section{Experiments}\label{sec:exp}
% 2026-11-03 (user, layout): Figure 3 defined at the start of Sec. 4 so that it prints on page 6 (Table 1 on page 5, Figure 4 on page 7).
% 2026-11-03 (user): the arXiv version (main_arxiv.tex defines \arxivversion) inputs it at the end of Sec. 3.5 instead, so that it prints on page 6 there too.
\ifdefined\arxivversion\else% Figure 3, input from 04_experiments.tex (anonymous main.pdf) or from 03_method.tex (arXiv, \arxivversion defined).
\begin{figure*}[t]
  \centering
  \includegraphics[width=\textwidth]{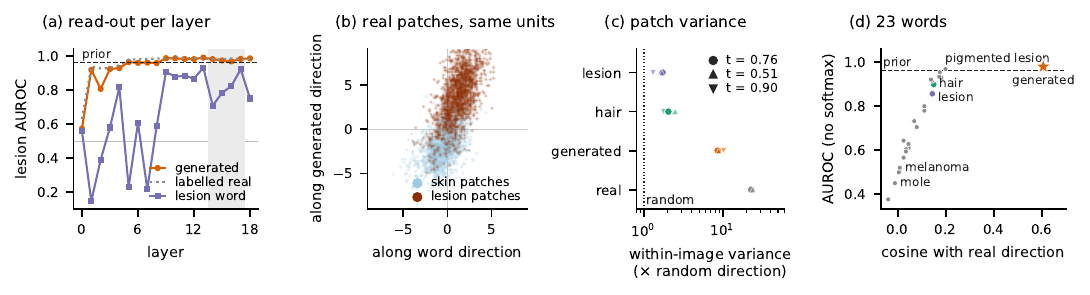}
  \caption{Why generated images work and the concept word fails (validation split; (b)--(d) at layers 14--17; dashed: mean-mask prior). (a) Lesion AUROC per double-stream layer of the word map, our map, and labelled real prototypes ($t=0.76$). (b) Patches minus their image mean along the unit word and generated directions (same units; 1{,}500 per class). (c) Within-image variance along each direction relative to a random one. (d) 23 words: cosine with the real skin-to-lesion direction against AUROC without softmax; star: generated direction.}
  \label{fig:why}
\end{figure*}
\fi

% 2026-10-18 (user): research questions pulled out of the Setup opening into their own subsection; 4.3-4.6 are headed by their RQ number.
\subsection{Research Questions}\label{sec:rq}

% 2026-11-03 (page compression): five questions as two sentences.
We first ask whether \methodname{} segments lesions better than label-free alternatives (\textbf{RQ1}, Section~\ref{sec:rq-compare}) and why generated images work as a query where the concept word fails (\textbf{RQ2}, Section~\ref{sec:rq-why}).
We then measure how much each module contributes (\textbf{RQ3}, Section~\ref{sec:rq-ablation}), how much the choices fixed on the validation split move the masks (\textbf{RQ4}, Section~\ref{sec:rq-sens}), and how the query and the threshold change a single mask (\textbf{RQ5}, Section~\ref{sec:rq-case}).

\subsection{Setup}\label{sec:setup}

% P1 (setup) [datasets]; the roadmap sentence became sec:rq (2026-10-18)
\textbf{Datasets.}
The test sets are the ISIC-2018 test set (1{,}000 images), PH2 (200 images, masks from a public mirror at $256{\times}256$), and 1{,}000 HAM10000 images, one per lesion, that share no image with ISIC-2018 lesion segmentation.
The labelled validation split holds 500 ISIC-2018 lesion-segmentation training images, disjoint from every test set.
As part of method development, it fixed the prompts, layers, noise level, and refinement fractions of \methodname{} once (Section~\ref{sec:rq-sens} varies them).
The refinement pool $\mathcal{U}$ of Section~\ref{sec:method-refine} is the same 500 images with their masks unread.
The prototypes $\mathbf{P}'$, $\mathbf{N}'$ and the cut-off $\tau$ are fixed before any test image is read, and each test image is segmented on its own.
% 2026-10-15 (user, option a): chest X-ray and polyp claims removed with their appendix section (additional datasets are not allowed in CVPR supplementary material).
% 2026-11-03 (review2, M6): run time from Supp. B (results/lineB/review/timing.json).
A test image takes 2.7~s on one A100 GPU (Supp.~B).

% P2 (setup) [model; baselines; nc_6]
\textbf{Model and Baselines.}
All methods that read FLUX use the unmodified FLUX.1-dev at $1024{\times}1024$, the layer set $\mathcal{L}=\{14,\dots,17\}$, noise level $t=0.76$, the caption ``a lesion'', and the guidance input of FLUX.1-dev set to 3.5; no weight is fine-tuned.
The baselines fall into four groups.
(1) \textbf{No image}: the mean-mask prior, the mean ground-truth mask of the other images of the same test set, which uses test labels.
(2) \textbf{Hand-designed for dermoscopy}: SLED, with four simpler methods in Supp.~E.
% 2026-11-03 (review2, M1): iSeg added (results/lineB/review2/c3_iseg_test.json).
(3) \textbf{External zero-shot methods}: MedCLIP-SAMv2, ADZUS, and iSeg, which iteratively refines Stable Diffusion cross-attention with self-attention.
ADZUS merges Stable Diffusion self-attention into regions and leaves the choice of the lesion region to a clinician, so we take the region nearest the image centre.
(4) \textbf{Word read-outs} score each patch against the concept word and are taken with public code from the comparison of ConceptAttention.
They are ConceptAttention at layers 14--17, as in its segmentation experiments, and at layers 9--13, FLUX cross-attention, DAAM on SD-XL, CLIP TextSpan, and seven CLIP explainability variants.
DINO self-attention, which uses no word, is listed with them.
% 2026-10-20 (user): moved from the method (Sec. 3.3 and 3.4).
ConceptAttention's word map and our map read the same attention outputs, so their comparison changes only what queries the features and how a patch is scored; the word map averages five noise draws in the full feature space, while \methodname{} uses one draw after a random projection (Supp.~A).
As a control, we also read our generated images and marked regions with BiomedCLIP, DINOv2, or CLIP features in place of FLUX's.
Baselines run from official or upstream code; Supp.~E lists the exceptions and settings.

% 2026-10-21 (user): Figure 3 is defined here, before Metrics, so that it prints on page 7 after Table 1 (page 6).
% 2026-10-18 (user): the mechanism of RQ2 as one figure; the per-layer curves (former fig:align) are its panels (a)-(b), and the competition panel was dropped as a repeat of fig:design (a) (results/lineB/figs/why.py).

% 2026-11-03 (user, layout): Figure 4 defined here so that it prints one page earlier (page 7).
% 2026-10-15 (user): module ablation and sensitivity merged into one row of four panels (results/lineB/figs/sensitivity.py).
\begin{figure*}[t]
  \centering
  \includegraphics[width=\textwidth]{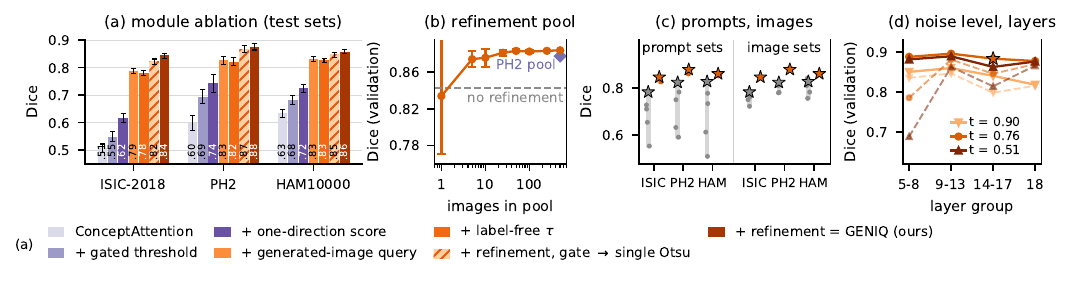}
  \caption{Design choices and sensitivity (stars: our setting). (a) From ConceptAttention to \methodname{}, one module at a time (test Dice, 95\% bootstrap intervals); hatched: the gate replaced by one per-image Otsu cut. (b) Validation Dice against unlabelled-pool size (mean $\pm$ s.d.\ over 20 draws). (c) Test Dice with generated images only (grey) and after refinement (orange) over five prompt sets and five generated image sets. (d) Validation Dice over three noise levels $t$ and four layer groups, with generated images only (dashed) and after refinement (solid).}
  \label{fig:design}
\end{figure*}

% P3 (setup) [metrics]
\textbf{Metrics.}
Masks are scored by Dice and IoU at $224{\times}224$, and maps by AUROC on the $32{\times}32$ patch grid.
IoU is $|M\cap Y|/|M\cup Y|$, which for one image equals $\mathrm{Dice}/(2-\mathrm{Dice})$, so the two rank masks alike per image and differ only when averaged, where IoU weighs poor overlaps more.
Confidence intervals are 95\% bootstrap intervals over test images, paired when two methods are compared on the same images.
Every map-producing baseline is thresholded by the rule of Section~\ref{sec:method-gap} with its own $\tau$ fitted on the labelled validation split (Supp.~I), which can only favour the baselines.
Mask-producing methods keep their own rule, and \methodname{} uses its label-free $\tau$.
Supp.~D gives the fitting grid and the histogram settings.
Table~S3 repeats the comparison under a single per-image Otsu threshold for every method.

\subsection{RQ1: Comparison with Label-Free Segmenters}\label{sec:rq-compare}

% P1 (RQ + setup)
\textbf{Setup.} We compare masks with the baselines of Section~\ref{sec:setup} on the three test sets (Table~\ref{tab:main}); Supp.~F shows the maps of every baseline on representative images.

% P2 (Findings) [c5 m6 nc_4]
% 2026-10-18 (user): findings state trends and comparisons; the numbers stay in the tables and figures they cite.
% \textbf{Findings.}
\textbf{Our masks outperform the pipeline hand-designed for dermoscopy in Dice and IoU on two test sets, and trail it on PH2.}
The paired differences on ISIC-2018 and HAM10000 have confidence intervals above zero, and on PH2 below zero (Tables~\ref{tab:main} and~S6).
% 2026-11-03 (review2, M5): strata from results/lineB/review2/b2_strata.json (Table S15).
SLED leads or ties on nevi, 80\% of PH2, and trails by up to 0.25 Dice on vignetted and non-nevus lesions (Supp.~H).
\textbf{Our masks outperform the external zero-shot methods and every word read-out on all three test sets.}
All paired Dice intervals against them exclude zero, as do the IoU intervals (ADZUS: Dice only; Tables~S6 and~S4).
This also holds under a single per-image Otsu threshold (Table~S3).
The closest, ConceptAttention at layers 9--13, trails ours by 0.07 to 0.14 Dice.
ADZUS stays below 0.55 Dice even with its regions chosen by test labels (Table~S4).
\textbf{Methods that define the target with real images instead of text do not reach better masks.}
Prototypes from ten labelled real images and centre-versus-border prototypes from the unlabelled validation images both trail ours by 0.001 to 0.011 Dice on every test set (Table~S9), while ours names the target by text.

\subsection{RQ2: Why the Concept Word Fails}\label{sec:rq-why}

% P1 (RQ + setup) [as_1]
\textbf{Setup.} We score the word with and without the softmax over the concept list and generated images on the three test sets, and compare directions among the attention outputs on the validation split.

% 2026-10-18 (user): RQ2 as a full argument, Figure fig:why next to fig:align; trends in the text, numbers in the figures and Table S12.
% Numbers: results/lineB/main/auroc_vs_prior.json, mech/module_ablation_test.json, mech/m1_m3.json, cvpr/v8_vocab.json, mech/v11_subspace_n{1,2,3}.json.
% \textbf{Findings.}
\textbf{Queried with the concept word, the attention outputs do not rank lesion patches above the mean-mask prior on any test set; queried with generated images, they do on all three.}
The word map at layers 14--17 is below the prior on ISIC-2018 and HAM10000 and level with it on PH2, while our map is above it on all three (Table~S12); on ISIC-2018 \texttt{hair} wins more than half of the lesion pixels.
On the validation split the word map is below the prior at every layer (Figure~\ref{fig:why}a).
\textbf{The features are not what fails: among the attention outputs, the direction from generated healthy skin to generated lesions lies close to the real one.}
From layer 9 on, the generated direction is far closer to the real skin-to-lesion direction than the \texttt{skin}-to-\texttt{lesion} direction, which is no closer than \texttt{skin}-to-\texttt{hair} (Figure~\ref{fig:why}d; per layer in Supp.~G), and generated prototypes rank lesion patches almost as well as labelled real ones (Figure~\ref{fig:why}a).
\textbf{Part of the word's failure is competition with the other words of the concept list.}
Scoring \texttt{lesion} against \texttt{skin} alone, without the softmax over the list, raises the word's AUROC on every test set, to about that of generated images on PH2 and HAM10000, but its Dice stays well below theirs on all three (Table~S12, Figure~\ref{fig:design}a).
The word direction also moves with the list, while the image tokens, and with them our query, do not attend to the concept tokens (Supp.~G).

\textbf{A better word does not close the gap.}
None of 23 words comes close to the generated direction in cosine with the real skin-to-lesion direction, and their AUROC does not follow their meaning: \texttt{melanoma} and \texttt{mole} are near chance, and \texttt{hair} ranks lesion patches better than \texttt{lesion} (Figure~\ref{fig:why}d).
The best word, \texttt{pigmented lesion}, edges past the prior but stays below the generated direction, and finding it takes labels.
\textbf{The patches of a real image barely vary along the word direction, and they do along the generated one.}
Along the \texttt{lesion} direction, the patch outputs of an image vary hardly more than along a random direction, as along \texttt{hair}.
Along the generated direction they vary several times more, at three noise levels (Figure~\ref{fig:why}c).
Placed along both directions in the same units, lesion and skin patches separate along the generated one and overlap along the word (Figure~\ref{fig:why}b).
The part of the word direction that lies where patches do vary is no more lesion-specific than that of \texttt{hair} (Table~S12).
The concept token's output thus differs from that of \texttt{skin} mostly where real patches do not differ.
The generated direction, a difference of image-token outputs, points where real lesions differ from skin.
These measurements are correlational and do not show why the model places the word there.
% 2026-11-03 (review2, M2): SD3-medium, results/lineB/review2/c1_sd3_test.json (Table S13).
\textbf{The same holds in a second diffusion transformer.}
In Stable Diffusion 3 medium, with its own generated images and only its layers and noise level chosen on the validation split, the word map reaches 0.25 to 0.45 Dice and ours 0.83 to 0.87, within 0.012 of FLUX (Table~S13).
\uline{The concept word competes with other words and points where real patches barely vary, while generated images point where lesions differ from skin.}

% REVISION R (2026-10-13): the small-lesion threshold analysis moved to Supp. app:binarisation (page budget); one sentence in 4.4.

% 2026-10-18 (user): the case study is the last figure. 2026-10-20 (user): RQ5, one row per phenomenon, not medians
% (results/lineB/review/case_rq5_assets.py; statistics case_rq5_stats.py -> case_rq5.json).
\begin{figure}[t]
  \centering
  \includegraphics[width=\linewidth]{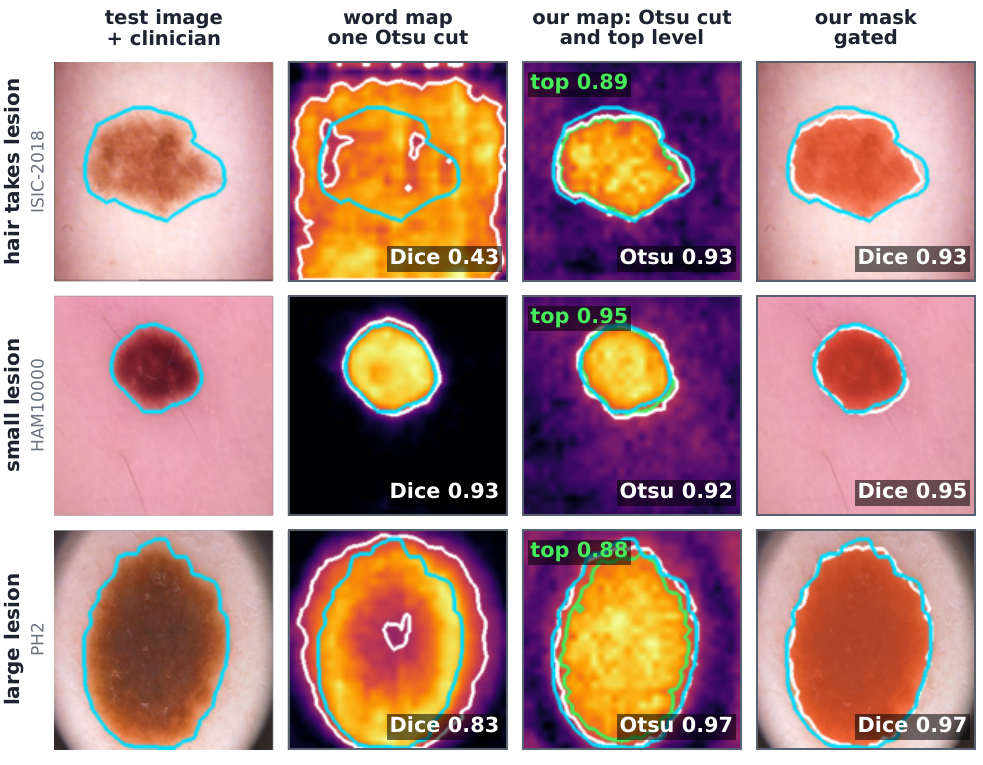}
  \caption{One change at a time on single test images (FLUX, layers 14--17). Columns 2 and 3 differ only in the query, both cut by one Otsu threshold (white). Columns 3 and 4 differ only in the threshold (green: top level of the three-level split; column 4: the gate's choice). Cyan: clinician's outline.}
  \label{fig:case}
\end{figure}

% 2026-10-19 (user): former RQ3 split into RQ3 (modules, fig:design a) and RQ4 (validation choices, fig:design b-d); layers are covered by fig:design d, so fig:why a is no longer cited here.
\subsection{RQ3: Module Contributions}\label{sec:rq-ablation}

\textbf{Setup.} Figure~\ref{fig:design}a turns ConceptAttention into \methodname{} one module at a time on the test sets; Tables~S7 and~S10 compare further choices.

% P2a (Findings, module ablation) [c18]
% \textbf{Findings.}
\textbf{Starting from ConceptAttention, each module of \methodname{} raises Dice on every test set; the generated-image query raises it most and the refinement second.}
The gated threshold and the one-direction score each add several Dice points, and setting $\tau$ without labels costs almost nothing (Figure~\ref{fig:design}a).
Within \methodname{}, a single Otsu cut in place of the gate (hatched bar) costs Dice mostly on the smallest lesions, whose predicted area the gate brings closer to the true one (Supp.~D); with the same inputs, OVDiff's many prototypes trail our single prototype difference by 0.07 to 0.09 validation Dice, even with healthy skin as background (Table~S10).

\subsection{RQ4: Sensitivity to the Validation Choices}\label{sec:rq-sens}

% P1 (RQ + setup) [user 2026-10-05: descriptive sensitivity]
\textbf{Setup.} Figure~\ref{fig:design}b--d varies one choice at a time.

% \textbf{Findings.}
\textbf{After refinement, the choices fixed on the validation split move the masks little.}
Refinement narrows the range of Dice over prompt sets, generated image sets, noise levels, and layer groups to a few points (Figure~\ref{fig:design}c--d), and averaging the prompt sets, so that no prompt is chosen with labels, keeps Dice where it is.
% 2026-11-03 (review2, M3): pool claims tightened, results/lineB/review2/b3_pool_tau.json (Table S14).
Ten unlabelled images give most of the gain (Figure~\ref{fig:design}b), and a pool from HAM10000 instead of ISIC-2018 costs 0.004 to 0.053 test Dice (Table~S14); concept lists and captions barely move our AUROC (Supp.~G).

% 2026-10-20 (user): case study as its own RQ, showing what the query and the threshold do to one mask; numbers are
% medians over each test set from results/lineB/review/case_rq5.json (ISIC-2018 / PH2 / HAM10000).
\subsection{RQ5: Case Study of the Query and the Threshold}\label{sec:rq-case}

\textbf{Setup.} Figure~\ref{fig:case} changes one thing at a time on single test images, one phenomenon per row on a typical image among those that show it (Supp.~F); the numbers below are medians over each test set.
The edge offset is the mean signed distance of a mask's edge from the clinician's, positive outside the lesion; one patch spans 7 pixels at $224{\times}224$.

% \textbf{Findings.}
\textbf{Queried with the concept word, the map can mark the skin around the lesion; queried with generated images, it marks the lesion.}
In the first row \texttt{hair} wins every lesion pixel, so the \texttt{lesion} map is low on the lesion and the Otsu cut takes the skin around it (Figure~\ref{fig:case}).
This is common rather than rare: \texttt{hair} wins more than a fifth of the lesion pixels on 67\%, 34\%, and 75\% of the test images, and under one Otsu cut the edge of the word mask lies 17, 23, and 4 pixels outside the clinician's, against 3, 2, and 0.5 for ours.
\textbf{On small lesions, the gate keeps the top level and removes the ring of spill-over that one Otsu cut keeps.}
In the second row the Otsu edge runs outside the lesion while the top level follows it.
On the smallest third of the lesions the gate keeps the top level on 76\% to 96\% of the images, and the edge offset falls from 8, 7, and 5 pixels to 3, 3, and 1, under one patch.
\textbf{On large lesions, the gate keeps one Otsu cut, since the top level alone covers only the dark core.}
In the third row the top level leaves out the lighter rim of the lesion.
On the largest third the top level alone would cover 57\% to 60\% of the lesion and reach Dice 0.72 to 0.75, while the gate keeps it on at most 3\% of the images and stays at 0.88 to 0.90.
The gate picks the better of its two cuts on 72\% to 79\% of all images and loses 0.014 to 0.017 Dice against it on average; Supp.~F shows a typical miss and what the refinement changes.
\uline{The query decides where the map lies, and the gate decides per image whether to cut at the lesion's edge or keep the whole lesion.}

% REVISION R (2026-10-13): refinement, text-only target, set of unlabelled images as a limitation.
% Tier 3 prose for the Conclusion (REFRAME T, task first, 2026-10-11): takeaway first, one paragraph; one method name;
% "generated images"; scopes out nc_1, nc_5, nc_8; main limitations here, the rest (incl. the TODO supervised-models
% sentence, nc_6) and the scope-outs in Supp. app:limitations (8-page budget, 2026-10-11).
\section{Conclusion}\label{sec:conclusion}

% 2026-10-18 (user): tightened for the page budget; the results sentence states the comparison without restating the tables.
% 2026-10-15 (user, option a): chest X-ray and polyp claims removed with their appendix section (additional datasets are not allowed in CVPR supplementary material).
% TODO (user, 2026-10-11): limitation on supervised models added without numbers; may be adjusted later.
We study skin lesion segmentation when the target is specified only by text, without annotated masks, hand-designed rules, model training, or additional models.
% 2026-11-03 (page compression): repeats of the first sentence and "in Dice and IoU" removed; the numbers stay.
A general text-to-image diffusion transformer can define the target without relying on a concept word alone: its generated images give a direction that better captures how real lesions differ from healthy skin. A few unlabelled dermoscopy images then adapt the lesion and healthy-skin prototypes to the target domain.
\methodname{} achieves Dice scores of 0.85–0.88 on three dermoscopy test sets. It outperforms all evaluated zero-shot methods and word-based read-outs on all three sets, and the hand-designed pipeline on two. It assumes position-invariant contrast (Assumption~2) and requires a few unlabelled target-domain images. Supervised segmenters achieve higher scores (Supp.~J).

% Citations are placeholders (user rule); enable with the first \cite.
{
    \small
    \bibliographystyle{ieeenat_fullname}
    \bibliography{reference}

@misc{zeng2023unsupervisedskinlesionsegmentation,
      title={Unsupervised Skin Lesion Segmentation via Structural Entropy Minimization on Multi-Scale Superpixel Graphs}, 
      author={Guangjie Zeng and Hao Peng and Angsheng Li and Zhiwei Liu and Chunyang Liu and Philip S. Yu and Lifang He},
      year={2023},
      eprint={2309.01899},
      archivePrefix={arXiv},
      primaryClass={cs.CV},
      url={https://arxiv.org/abs/2309.01899}, 
}

@misc{ramella2021saliencybasedsegmentationdermoscopicimages,
      title={Saliency-based segmentation of dermoscopic images using color information}, 
      author={Giuliana Ramella},
      year={2021},
      eprint={2011.13179},
      archivePrefix={arXiv},
      primaryClass={eess.IV},
      url={https://arxiv.org/abs/2011.13179}, 
}

@misc{helbling2025conceptattentiondiffusiontransformerslearn,
      title={ConceptAttention: Diffusion Transformers Learn Highly Interpretable Features}, 
      author={Alec Helbling and Tuna Han Salih Meral and Ben Hoover and Pinar Yanardag and Duen Horng Chau},
      year={2025},
      eprint={2502.04320},
      archivePrefix={arXiv},
      primaryClass={cs.CV},
      url={https://arxiv.org/abs/2502.04320}, 
}

@misc{tang2022daaminterpretingstablediffusion,
      title={What the DAAM: Interpreting Stable Diffusion Using Cross Attention}, 
      author={Raphael Tang and Linqing Liu and Akshat Pandey and Zhiying Jiang and Gefei Yang and Karun Kumar and Pontus Stenetorp and Jimmy Lin and Ferhan Ture},
      year={2022},
      eprint={2210.04885},
      archivePrefix={arXiv},
      primaryClass={cs.CV},
      url={https://arxiv.org/abs/2210.04885}, 
}

@misc{koleilat2025medclipsamv2universaltextdrivenmedical,
      title={MedCLIP-SAMv2: Towards Universal Text-Driven Medical Image Segmentation}, 
      author={Taha Koleilat and Hojat Asgariandehkordi and Hassan Rivaz and Yiming Xiao},
      year={2025},
      eprint={2409.19483},
      archivePrefix={arXiv},
      primaryClass={cs.CV},
      url={https://arxiv.org/abs/2409.19483}, 
}

@misc{hamrani2025selfattentiondiffusionmodelszeroshot,
      title={Self-Attention Diffusion Models for Zero-Shot Biomedical Image Segmentation: Unlocking New Frontiers in Medical Imaging}, 
      author={Abderrachid Hamrani and Anuradha Godavarty},
      year={2025},
      eprint={2503.18170},
      archivePrefix={arXiv},
      primaryClass={cs.CV},
      url={https://arxiv.org/abs/2503.18170}, 
}

@misc{kirillov2023segment,
      title={Segment Anything}, 
      author={Alexander Kirillov and Eric Mintun and Nikhila Ravi and Hanzi Mao and Chloe Rolland and Laura Gustafson and Tete Xiao and Spencer Whitehead and Alexander C. Berg and Wan-Yen Lo and Piotr Dollár and Ross Girshick},
      year={2023},
      eprint={2304.02643},
      archivePrefix={arXiv},
      primaryClass={cs.CV},
      url={https://arxiv.org/abs/2304.02643}, 
}

@misc{tian2024diffuseattendsegmentunsupervised,
      title={Diffuse, Attend, and Segment: Unsupervised Zero-Shot Segmentation using Stable Diffusion}, 
      author={Junjiao Tian and Lavisha Aggarwal and Andrea Colaco and Zsolt Kira and Mar Gonzalez-Franco},
      year={2024},
      eprint={2308.12469},
      archivePrefix={arXiv},
      primaryClass={cs.CV},
      url={https://arxiv.org/abs/2308.12469}, 
}

@misc{kim2025seg4diffunveilingopenvocabularysegmentation,
      title={Seg4Diff: Unveiling Open-Vocabulary Segmentation in Text-to-Image Diffusion Transformers}, 
      author={Chaehyun Kim and Heeseong Shin and Eunbeen Hong and Heeji Yoon and Anurag Arnab and Paul Hongsuck Seo and Sunghwan Hong and Seungryong Kim},
      year={2025},
      eprint={2509.18096},
      archivePrefix={arXiv},
      primaryClass={cs.CV},
      url={https://arxiv.org/abs/2509.18096}, 
}

@misc{zhang2026anchordifftrainingfreeconceptgrounding,
      title={AnchorDiff: Training-Free Concept Grounding for MM-DiTs via Anchor-Based Graph Propagation}, 
      author={Jian Zhang and Zhijun Zhang},
      year={2026},
      eprint={2605.26460},
      archivePrefix={arXiv},
      primaryClass={cs.CV},
      url={https://arxiv.org/abs/2605.26460}, 
}

@misc{karazija2024diffusionmodelsopenvocabularysegmentation,
      title={Diffusion Models for Open-Vocabulary Segmentation}, 
      author={Laurynas Karazija and Iro Laina and Andrea Vedaldi and Christian Rupprecht},
      year={2024},
      eprint={2306.09316},
      archivePrefix={arXiv},
      primaryClass={cs.CV},
      url={https://arxiv.org/abs/2306.09316}, 
}

@misc{barsellotti2024trainingfreeopenvocabularysegmentationoffline,
      title={Training-Free Open-Vocabulary Segmentation with Offline Diffusion-Augmented Prototype Generation}, 
      author={Luca Barsellotti and Roberto Amoroso and Marcella Cornia and Lorenzo Baraldi and Rita Cucchiara},
      year={2024},
      eprint={2404.06542},
      archivePrefix={arXiv},
      primaryClass={cs.CV},
      url={https://arxiv.org/abs/2404.06542}, 
}

@misc{labs2025flux1kontextflowmatching,
      title={FLUX.1 Kontext: Flow Matching for In-Context Image Generation and Editing in Latent Space},
      author={Black Forest Labs and Stephen Batifol and Andreas Blattmann and Frederic Boesel and Saksham Consul and Cyril Diagne and Tim Dockhorn and Jack English and Zion English and Patrick Esser and Sumith Kulal and Kyle Lacey and Yam Levi and Cheng Li and Dominik Lorenz and Jonas Müller and Dustin Podell and Robin Rombach and Harry Saini and Axel Sauer and Luke Smith},
      year={2025},
      eprint={2506.15742},
      archivePrefix={arXiv},
      primaryClass={cs.GR},
      url={https://arxiv.org/abs/2506.15742},
}

@misc{flux2024,
    author={Black Forest Labs},
    title={FLUX},
    year={2024},
    howpublished={\url{https://github.com/black-forest-labs/flux}},
}

@misc{jiang2025advancinggeneralizabletumorsegmentation,
      title={Advancing Generalizable Tumor Segmentation with Anomaly-Aware Open-Vocabulary Attention Maps and Frozen Foundation Diffusion Models}, 
      author={Yankai Jiang and Peng Zhang and Donglin Yang and Yuan Tian and Hai Lin and Xiaosong Wang},
      year={2025},
      eprint={2505.02753},
      archivePrefix={arXiv},
      primaryClass={cs.CV},
      url={https://arxiv.org/abs/2505.02753}, 
}

@misc{vilouras2025zeroshotmedicalphrasegrounding,
      title={Zero-Shot Medical Phrase Grounding with Off-the-shelf Diffusion Models}, 
      author={Konstantinos Vilouras and Pedro Sanchez and Alison Q. O'Neil and Sotirios A. Tsaftaris},
      year={2025},
      eprint={2404.12920},
      archivePrefix={arXiv},
      primaryClass={cs.CV},
      url={https://arxiv.org/abs/2404.12920}, 
}

@misc{tang2025prototypebasedimagepromptingweakly,
      title={Prototype-Based Image Prompting for Weakly Supervised Histopathological Image Segmentation}, 
      author={Qingchen Tang and Lei Fan and Maurice Pagnucco and Yang Song},
      year={2025},
      eprint={2503.12068},
      archivePrefix={arXiv},
      primaryClass={cs.CV},
      url={https://arxiv.org/abs/2503.12068}, 
}

@misc{wang2024diffusionmodelsecretlytrainingfree,
      title={Diffusion Model is Secretly a Training-free Open Vocabulary Semantic Segmenter}, 
      author={Jinglong Wang and Xiawei Li and Jing Zhang and Qingyuan Xu and Qin Zhou and Qian Yu and Lu Sheng and Dong Xu},
      year={2024},
      eprint={2309.02773},
      archivePrefix={arXiv},
      primaryClass={cs.CV},
      url={https://arxiv.org/abs/2309.02773}, 
}

@misc{kukleva2026refamattentionmagnetszeroshot,
      title={RefAM: Attention Magnets for Zero-Shot Referral Segmentation}, 
      author={Anna Kukleva and Enis Simsar and Alessio Tonioni and Muhammad Ferjad Naeem and Federico Tombari and Jan Eric Lenssen and Bernt Schiele},
      year={2026},
      eprint={2509.22650},
      archivePrefix={arXiv},
      primaryClass={cs.CV},
      url={https://arxiv.org/abs/2509.22650}, 
}

@misc{chakraborty2026mavisegmanifoldpropagationvisual,
      title={MAVISEG: Manifold Propagation and Visual Prototypes for Zero-Shot Open-Vocabulary Segmentation in Diffusion Transformers}, 
      author={Rajatsubhra Chakraborty and Xujun Che and Ritabrata Chakraborty and Xi Niu and Depeng Xu},
      year={2026},
      eprint={2608.05878},
      archivePrefix={arXiv},
      primaryClass={cs.CV},
      url={https://arxiv.org/abs/2608.05878}, 
}

@misc{jun2026interpretablemotionattentivemapsspatiotemporally,
      title={Interpretable Motion-Attentive Maps: Spatio-Temporally Localizing Concepts in Video Diffusion Transformers}, 
      author={Youngjun Jun and Seil Kang and Woojung Han and Seong Jae Hwang},
      year={2026},
      eprint={2603.02919},
      archivePrefix={arXiv},
      primaryClass={cs.CV},
      url={https://arxiv.org/abs/2603.02919}, 
}

@InProceedings{Barsellotti_2024_WACV,
    author    = {Barsellotti, Luca and Amoroso, Roberto and Baraldi, Lorenzo and Cucchiara, Rita},
    title     = {FOSSIL: Free Open-Vocabulary Semantic Segmentation Through Synthetic References Retrieval},
    booktitle = {Proceedings of the IEEE/CVF Winter Conference on Applications of Computer Vision (WACV)},
    month     = {January},
    year      = {2024},
    pages     = {1464-1473}
}

@misc{ma2026perceptualanchoringprototypeguidedtext,
      title={Perceptual Anchoring: Prototype-Guided Text Calibration for Training-free Open-Vocabulary Semantic Segmentation}, 
      author={Wanli Ma and Jiangwen Lu and Qinmu Peng and Xinge You},
      year={2026},
      eprint={2608.03991},
      archivePrefix={arXiv},
      primaryClass={cs.CV},
      url={https://arxiv.org/abs/2608.03991}, 
}

@misc{wu2024diffumasksynthesizingimagespixellevel,
      title={DiffuMask: Synthesizing Images with Pixel-level Annotations for Semantic Segmentation Using Diffusion Models}, 
      author={Weijia Wu and Yuzhong Zhao and Mike Zheng Shou and Hong Zhou and Chunhua Shen},
      year={2024},
      eprint={2303.11681},
      archivePrefix={arXiv},
      primaryClass={cs.CV},
      url={https://arxiv.org/abs/2303.11681}, 
}

@misc{mao2025medsegfactorytextguidedgenerationmedical,
      title={MedSegFactory: Text-Guided Generation of Medical Image-Mask Pairs}, 
      author={Jiawei Mao and Yuhan Wang and Yucheng Tang and Daguang Xu and Kang Wang and Yang Yang and Zongwei Zhou and Yuyin Zhou},
      year={2025},
      eprint={2504.06897},
      archivePrefix={arXiv},
      primaryClass={cs.CV},
      url={https://arxiv.org/abs/2504.06897}, 
}

@ARTICLE{Otsu4310076,
  author={Otsu, Nobuyuki},
  journal={IEEE Transactions on Systems, Man, and Cybernetics}, 
  title={A Threshold Selection Method from Gray-Level Histograms}, 
  year={1979},
  volume={9},
  number={1},
  pages={62-66},
  doi={10.1109/TSMC.1979.4310076}}

@misc{snell2017prototypicalnetworksfewshotlearning,
      title={Prototypical Networks for Few-shot Learning}, 
      author={Jake Snell and Kevin Swersky and Richard S. Zemel},
      year={2017},
      eprint={1703.05175},
      archivePrefix={arXiv},
      primaryClass={cs.LG},
      url={https://arxiv.org/abs/1703.05175}, 
}
}

\end{document}